%% file: SSSCAM-arxiv2.tex
 \documentclass{article}

\input{macros.tex}

\usepackage{fullpage}

\begin{document}

\title{SSSC-AM: A Unified Framework for Video Co-Segmentation by Structured Sparse Subspace Clustering with Appearance and Motion Features}

\author{ Junlin Yao\thanks{\'Ecole Polytechnique, Palaiseau, France.} 
        \and~Frank~Nielsen\thanks{\'Ecole Polytechnique, Palaiseau, France. Corresponding author: Frank Nielsen (email: \url{Frank.Nielsen@acm.org})}
}

\maketitle

\begin{abstract}
Video co-segmentation  refers to the task of jointly segmenting common objects appearing in a given group of videos. 
In practice, high-dimensional data such as videos can be conceptually thought as being drawn from a union of subspaces corresponding to categories rather than from a  smooth manifold. 
Therefore, segmenting data into respective subspaces --- subspace clustering --- finds widespread applications in computer vision, including co-segmentation. 
State-of-the-art methods via subspace clustering seek to solve the problem in two steps:
 First, an affinity matrix is built from data, with appearance features or motion patterns. 
Second, the data are segmented by applying spectral clustering to the affinity matrix. 
However, this process is insufficient to obtain an optimal solution since it does not take into account the {\em interdependence} of the affinity matrix with the segmentation. 
In this work, we present a novel unified video co-segmentation framework inspired by the recent Structured Sparse Subspace Clustering ($\mathrm{S^{3}C}$) based on the {\em self-expressiveness} model. 
Our method yields more consistent segmentation results. 
In order to improve the detectability of motion features with missing trajectories due to occlusion or tracked points moving out of frames, 
we add an extra-dimensional signature to the motion trajectories. Moreover, we reformulate the $\mathrm{S^{3}C}$ algorithm by adding the affine subspace constraint in order to make it more suitable to segment rigid motions lying in affine subspaces of dimension at most $3$. 
Our experiments on MOViCS dataset show that our framework achieves the highest overall performance among baseline algorithms and demonstrate its robustness to heavy noise.

\end{abstract}


\section{Introduction, prior work and contributions}
\label{sec:intro}
Over the past several decades, we have witnessed an explosion in the availability of video data.
Just to give a figure, let us say that the amount of new videos uploaded to {\sc YouTube} every minute is estimated to be $300$ hours!
Segmenting videos into multiple spatio-temporal areas, and extracting useful information effectively and computationally efficiently from them becomes a crucial issue and has broad applications, such as $3\mathrm{D}$ reconstruction, action recognition, etc. 
Various methods have been proposed to deal with this challenging video segmentation problem. 
For example, subspace clustering based methods~\cite{Vidal2010} seek to discover low-dimensional representation of high-dimensional data points.

Co-segmentation~\cite{Guo2014,Wang2015,Arm2010} consists in segmenting simultaneously multiple images or videos into several subregions denoting common objects.
Thus co-segmentation searches for objects jointly appearing in videos, and provides additional information in the absence of supervisory information~\cite{Arm2010}.

\subsection{Previous work}
\label{subsec:previous}
\subsubsection{Subspace clustering}
\label{subsubsec:sc}

Many subspace clustering algorithms have already been proposed, including algebraic algorithms~\cite{Vidal}, iterative methods~\cite{Agarwal}, 
statistical methods~\cite{Fischler}, and spectral clustering based methods \cite{Luxburg2007,Li2015,Soltanolkotabi2012}. Among these algorithms,
 the Structured Sparse Subspace Clustering ($\mathrm{S^{3}C}$ for short) algorithm~\cite{Li2015a} outperforms all others due to its framework promoting coherence between the affinity matrix and segmentation, as well as its robustness with respect to noise and outliers. However, $\mathrm{S^{3}C}$  only considers linear subspaces while it is frequent to encounter data lying in a union of {\em affine} rather than linear subspaces for dynamic data-sets. 
For instance, the motion segmentation problem consists in clustering data drawn from $3$-dimensional affine subspaces. Though there exists a naive way that data coming from affine subspaces can be treated as if they lie in linear subspaces, this method may be potentially unable to distinguish subspaces from each other~\cite{Elhamifar}. Since video co-segmentation involves motion segmentation and thus affine subspaces, it is necessary for the subspace clustering method to have the ability to deal with data points lying in affine structures.

\subsubsection{Video co-segmentation}
\label{subsubsec:coseg}
Several methods, such as ~\cite{Chiu2013,Wang2014}, have been proposed to tackle the video co-segmentation problem. Both of the methods, similar to our framework, seek to jointly segment objects in video sequences in the absence of supervisory information. In~\cite{Chiu2013}, Chiu et al. formulate a generative multi-video model to enable multi-class video co-segmentation. In the presence of noisy motion information, such as objects moving together or existence of similar motions, its segmentation performance degrades quickly. In~\cite{Wang2014}, Wang et al. propose a co-segmentation algorithm given facts exploited by subspace clustering based method called \textit{Low-Rank Representation} (LRR). Both their method and ours conceptually view video data as being drawn from a union of subspaces. Compared to their assumption of background/foreground binary classification, our framework assumes a multi-class structure of videos and yields multi-class segmentation results as in~\cite{Chiu2013}. Also, their algorithm suffers from limitations that it is unable to tackle defocus blur, noise and outliers. 
Inspired by those recent works, our framework seeks to better cope with videos corrupted by noise and outliers, as demonstrated by our experiments, where ~\cite{Chiu2013,Wang2014} serve as a benchmark.

\subsection{Paper contributions}
\label{subsec:contributions}
In this paper, we propose a {\em unified} framework for video co-segmentation that enables to obtain {\em better coherent segmentation} results. 
We summarize our contributions as follows:
First, we improve the detectability of motion trajectories corrupted by missing data using extra-dimensional signatures. 
Second, in order to deal with motion features lying in $3$-dimensional affine subspaces, we reformulate the structured sparse subspace clustering by adding the affine subspace constraint, and solve this novel optimization problem efficiently using the Alternating Direction Method of Multipliers (ADMM) method. 
By integrating this subspace clustering approach, we propose a unified optimization framework, dubbed Structured Sparse Subspace Clustering with Appearance and Motion Features ($\mathrm{S^{3}C-AM}$).

\subsection{Paper organization}
\label{subsec:org}
In Section \ref{sec:sc}, we introduce the subspace clustering problem and the $\mathrm{S^{3}C}$ algorithm for clustering data points lying in a union of linear subspaces. In Section \ref{sec:framework}, we motivate and formulate the unified framework for video co-segmentation. In Section \ref{sec:experiments}, we study experimentally the effectiveness our framework   on real video data sets. In Section \ref{sec:future}, we discuss on some possible extensions of our work. Finally, Section \ref{sec:conclusion} concludes the paper.
The notations used throughout the paper are concisely recalled in Appendix~A.

\section{Subspace Clustering}
\label{sec:sc}

\subsection{Overview}
\begin{figure}[t]
\centering
\includegraphics[width=\columnwidth]{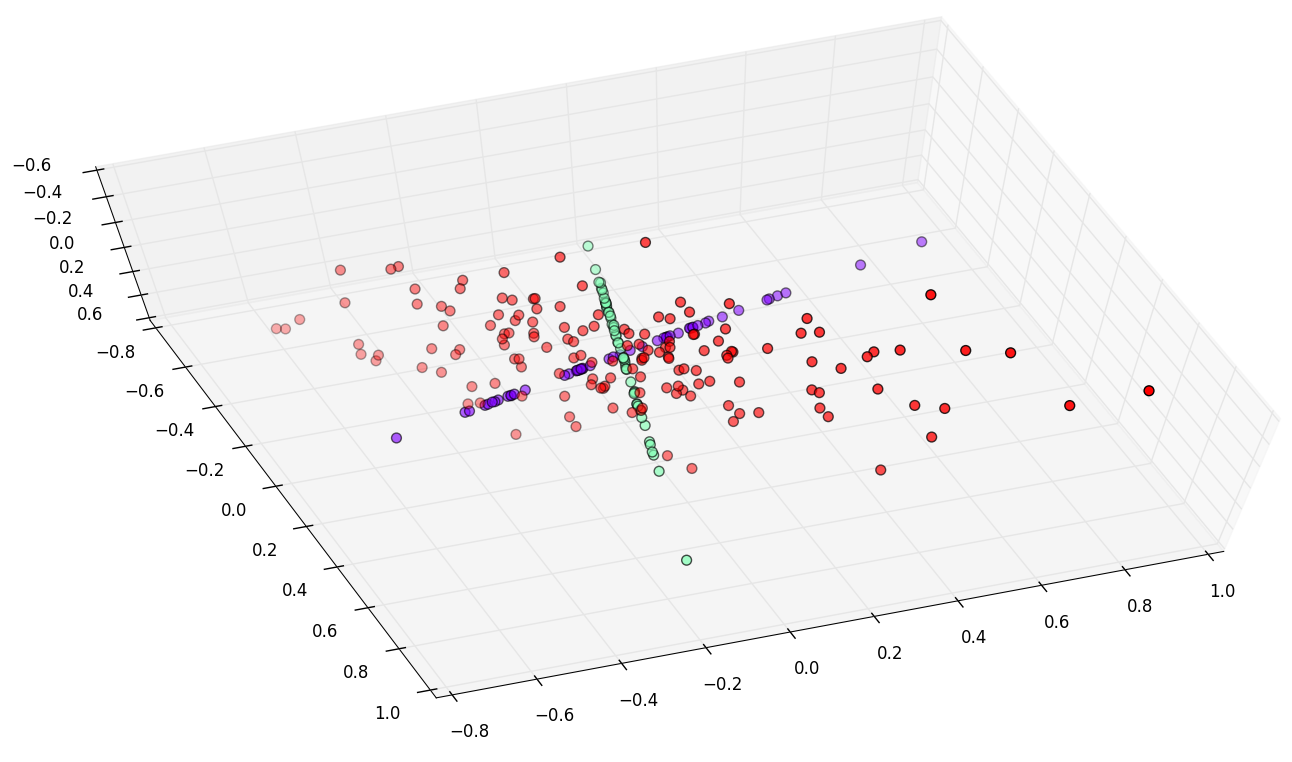}
\caption{Three subspaces in $\mathbb{R}^{3}$: the first and second group of $50$ data points belong to two one-dimensional subspaces, and the remaining $150$ data points belong  to a two-dimensional subspace.}
\label{fig:subspaces}
\end{figure}

\emph{Subspace Clustering} consists in finding low-dimensional representation of high-dimensional data. For example, given data points drawn from a three-dimensional space, we recover their subspace structures via subspace clustering as depicted in Figure \ref{fig:subspaces}.

The subspace clustering problem can be framed mathematically as follows.
\theoremstyle{plain} \newtheorem{problem}{Problem}
\begin {problem}[Subspace clustering] \label{pb:sc}
Let $X\in\mathbb{R}^{n\times{N}}$ be a real-valued matrix whose columns are $N$ data points drawn from a union of $M$ subspaces $S_j$ of $\mathbb{R}^n$, of dimensions $d_j<\min{\{n,N\}}$, for $j=1,\ldots,M$. Given $X$, the subspace clustering seeks to find: a) the subspace to which each data point belongs to. That is, the segmentation of the data points; b) the number of subspaces and their intrinsic dimensions.
\end{problem}

Note that finding the respective basis of subspaces can be obtained as a by-product of solving the subspace clustering problem.
Recent approaches to solving  problem~\ref{pb:sc} are based on the \emph{self-expressiveness model}~\cite{Soltanolkotabi2012,Li2015a}, which states that each data point in a union of subspaces can be expressed as a linear or an affine combination of other points. 
When data points lie in a linear subspace $S_{j}$ of dimension $d_{j}$, each data point can be written as a linear combination of $d_{j}$ other points from $S_{j}$. Similarly, when data points lie in an affine subspace $S_{j}$ of dimension $d_{j}$, each data point can be written as an affine combination of $d_{j}+1$ other points from $S_{j}$. The combination can be represented using matrix notation by $X=XZ$, where $X$ denotes the {\em data matrix} and $Z$ the {\em matrix of coefficients}. 

Let us first consider the problem ~\ref{pb:sc} in the case of linear subspaces. Although this combination is in general not unique, a {\em sparse representation} of a data point ideally corresponds to a combination of a few points belonging to its own subspace~\cite{Elhamifar}. By solving a global sparse optimization problem, and obtaining the matrix of coefficients, the spectral clustering technique is then applied to infer the clustering of data points, whose theoretical bound can be guaranteed by the normalized cut~\cite{malik}. 
In addition, the number of the nonzero elements in the sparse representation of a data point ideally corresponds to the dimension of its subspace. 
As for the number of subspaces, it can be found either by analyzing the eigenspectrum of the Laplacian matrix of the graph constructed by $Z$~\cite{Luxburg2007}, or by model selection techniques~\cite{Brox2010}. Finally, the respective basis of subspaces can be readily obtained. Therefore, with the sparse representation framework based on self-expressiveness, we can solve thoroughly the subspace clustering problem~\ref{pb:sc}. Compared to local spectral clustering-based algorithms, we can furthermore overcome challenges such as dealing with points near the intersection of subspaces.

When it comes to the case of affine subspaces, the property that a data point can be represented sparsely by a few points from its own subspace becomes more subtle. \cite{Li2015On-Sufficient-C} introduces situations where the property fails to be met when a grouping effect for interior points is present. Nonetheless, our experiments show that the sparse self-expressiveness in the case of affine subspaces is satisfied in our context.

\subsection{Structured sparse subspace clustering ($\mathrm{S^{3}C}$)}

State-of-the-art subspace clustering methods based on the self-expressiveness model follow a two-step approach to solve the subspace clustering problem \ref{pb:sc}: learning an affinity matrix from data in the first step and segmenting data based on the affinity matrix in the second step. This two-step approach is generally suboptimal since it does not consider the interdependence of the affinity with the segmentation. 
In~\cite{Li2015a}, Li et al. seek to combine these two steps by building a unified optimization framework, dubbed {\em Structured Sparse Subspace Clustering} (SSSC or $\mathrm{S^3C}$) in which obtaining affinity matrix and segmentation of data are merged together. 
More specifically, this approach considers the $\mathrm{S^3C}$ problem as follows:
\begin{eqnarray}{l}\label{eqn:opt_pb}
\min_{Z,E,Q}\norm{Z}_{1,Q}+\lambda\norm{E}_\ell		\nonumber \\
\mathrm{s.t.}\; X=XZ+E,\; \mathrm{diag}(Z)=\bm{0},\; Q\in \mathcal{Q}.	 
\end{eqnarray}

where $X$ is the $n\times N$ data matrix and $Z$ the coefficient matrix. The norm $\norm{\cdot}_\ell$ on the error term $E$ depends upon the prior knowledge about the pattern of noise. $Q=[\mathbf{q_{1}}, \cdots, \mathbf{q_{M}}]$ refers to an $N\times M$ {\em binary segmentation matrix} indicating the membership of each data point to each subspace. That is, $q_{i,j}=1$ if the $i$-th column of $X$ lies in subspace $S_{j}$ and $q_{i,j}=0$ otherwise. The space of segmentation matrices is defined as $\mathcal{Q}=\{Q\in \{0,1\}^{N\times M}:Q\mathbf{1}=\mathbf{1}\: \mathrm{and\:rank}(Q)=M\}$. $\lambda$ and $\alpha$ are two trade-off parameters. The \textit{subspace structured} $\ell_1$ norm of $Z$ is defined as follows:

\begin{eqnarray}{rCl}
\label{eqn:ss_norm}
\norm{Z}_{1,Q} & \stackrel{\cdot}{=} & \norm{Z}_1+\alpha\norm{\Theta\odot{Z}}_1	\nonumber \\
&=&\sum_{i,j}\vert{Z_{i,j}}\vert \left(1+\frac{\alpha}{2}\norm{\bm{q}^{(i)}-\bm{q}^{(j)}}^2\right)	 
\end{eqnarray}
where $\Theta_{i,j} = \frac{1}{2}\norm{\bm{q}^{(i)}-\bm{q}^{(j)}}^2$ with $\bm{q}^{(i)}$ and $\bm{q}^{(i)}$ the $i$-th and $j$-th row of matrix $Q$, respectively.

The optimization problem~\eqref{eqn:opt_pb} ensures a sparse representation of data points by means of the  $\ell_1$ norm. The constraint $X=XZ+E$ seeks to capture the self-expressiveness while $\mathrm{diag}(Z)=\bm{0}$ prevents data points representing themselves. The introduction of \eqref{eqn:ss_norm} merges the affinity and segmentation together. 
By formulating the subspace clustering problem in this way, $\mathrm{S^{3}C}$ is able to yield more coherent clustering results.
We build our unified framework for video co-segmentation based on $\mathrm{S^{3}C}$.

\section{A unified framework for video co-segmentation}
\label{sec:framework}
In this section, we introduce our unified framework for video co-segmentation with \emph{appearance} and \emph{motion features}. Appearance feature and motion feature are both crucial in object segmentation and recognition. On one hand, the appearance feature captures color or texture information of objects. It emphasizes the salience of an object by which it stands out relative to its neighbors.
 On the other hand, the motion feature shows how objects move. By exploiting together the appearance and motion information, we simulate thus how human beings recognize and differentiate objects. We given an overview of our  framework   as follows:
  
First, we build temporal superpixels in the preprocessing stage  in order to reduce computational costs.
We extract motion and appearance features from those superpixels. 
Furthermore, we improve the detectability of motion features by adding an {\em extra signature} to the conventional motion trajectory matrix. Then, we formulate the unified optimization framework with the additional affine subspace constraint, and solve our global optimization problem efficiently using the Alternating Direction Method of Multipliers (ADMM) method. Finally, we construct a unified affinity matrix from the optimal solution of the optimization problem, and obtain segmentation results by applying spectral clustering to the affinity matrix.

\subsection{Preprocessing: Temporal superpixels}

In order to reduce the computational complexity, we represent videos using Temporal SuperPixels (TSPs)~\cite{Chang} which help build superpixel-wise spatio-temporal correspondences. 
TSPs are different from common over-segmentation techniques because those superpixel representations maintain the most salient features of an image. In addition, we integrate SIFT flow~\cite{Liu2011,Wang2014} into the TSP framework so as to improve the robustness of correspondences. 
We then extract motion and appearance features at the level of TSPs.

\subsection{Appearance features}
We choose the appearance feature to be $\mathrm{10D}$ color features in HSV color space. 
Gaussian mixture model and discretization are then applied to the color features to obtain the color histogram~\cite{Zhang2012}.

\subsection{Motion feature using an extra-dimensional signature}
In order to capture motion information of objects, we need a function called a \emph{camera model} to map the three-dimensional world onto a two-dimensional \emph{image plane}. In the case of imaging objects far away from a camera, small differences in depth, or the \emph{perspective effect}, 
becomes less apparent~\cite{Poling}. The perspective projection can thus be approximated by an {\em affine projection} with the \emph{affine camera model}. 
Thus we construct the motion feature under the affine projection model to fit the affine subspace clustering method.

Traditionally~\cite{Vidal2010}, motion feature matrix is represented by \eqref{eqn:motion_feature_original}.
\begin{eqnarray}{l}
\label{eqn:motion_feature_original}
  M = \left[\begin{array}{ccc}
    x_{11} & \ldots & x_{1N}\\
    \vdots &  & \vdots\\
    x_{F1} & \ldots & x_{FN}
  \end{array}\right]_{2 F \times N}  
\end{eqnarray}
where $\{ x_{\tmop{fj}} \in \mathbbm{R}^2 \}^{f = 1, \ldots, F}_{j = 1,
\ldots, N}$ denotes the $2\mathD$ projection of $N$ $3\mathD$ points
on rigidly moving objects onto $F$ frames of a moving camera. In our case of using superpixels,  $x_{\tmop{fj}}$ refers to the center of the intersection of $j$-th TSP and $f$-th frame. However, in real-world problems, there often exist missing trajectories caused by tracked points moving into or out of frames, or by occlusion. Conventionally, nearest motion values in the same data column are padded to the missing entries~\cite{Wang2014}. While this simple treatment sometimes results in acceptable segmentation, generally it reduces the detectability of motion trajectories, making subspace clustering method unable to perform motion segmentation. To overcome this limitation, in addition to padding motion values, we propose a new way to form motion feature matrix by adding an \emph{extra-dimension signature} so that $\{ x_{\tmop{fj}} \in \mathbbm{R}^3 \}^{f = 1, \ldots, F}_{j = 1, \ldots, \Nu}$ with $x_{\tmop{fj}} = (x\;y\; s)^{\top}$. $s$ denotes the signature added to the original motion trajectories. Different values of $s$ indicate different {\em status} of motion trajectories, such as missing trajectories caused by newly appearing points or by dead points, which can be readily inferred from indices of TSPs . Newly appearing TSPs have indices that do not exist in previous frames while indices assigned to dead TSPs disappear starting from certain frame. The values of the signature, $s_{i},\; i=1,2,3$ are specified so as to encode the status of motion trajectories and sufficiently distinguish one from another. Here, we set $s_{i},\; i=1,2,3$ to be $1,3,5$, respectively, as shown in Table~\ref{tab:mapping}.

\begin{table}[htbp]
\centering
\begin{tabular}{|c|c|}
\hline
Value & Status of motion trajectories \\
\hline
$s_{1}=1$ & Original complete trajectories \\
$s_{2}=3$ & Missing trajectories caused by new points \\
$s_{3}=5$ & Missing trajectories caused by dead points \\
\hline
\end{tabular}
\caption{Meaning of signature $s$ to status of motion trajectories: Depending on newly appearing or disappearing feature points, we encode the event using an extra  dummy dimension to get better and more consistent subspace clustering results.}
\label{tab:mapping}
\end{table}

\begin{figure}[t]
\centering
\subfloat[Original frame]{
	\label{fig:original_frame_1}
	\includegraphics[width=2.3cm]{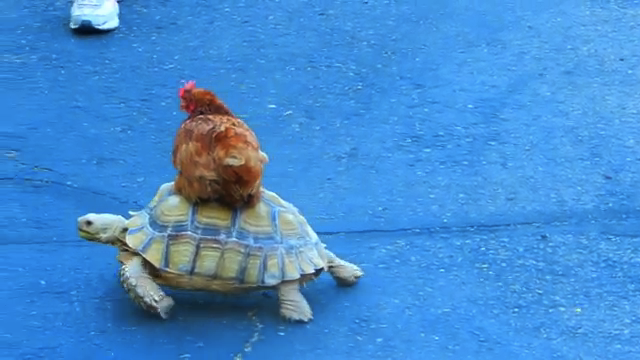}
}
\hspace{0.02\linewidth}
\subfloat[With signature]{
	\label{fig:with_signature}
	\includegraphics[width=2.3cm]{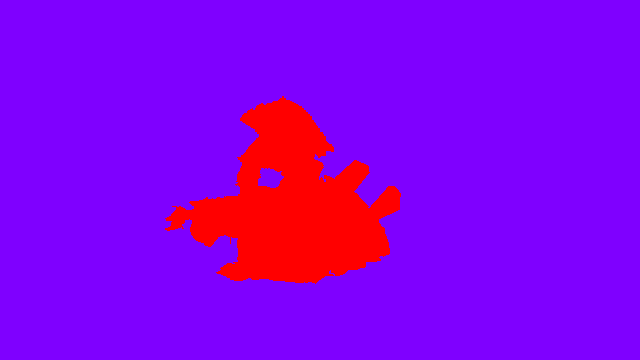}
}
\hspace{0.02\linewidth}
\subfloat[Without signature]{
	\label{fig:without_signature}
	\includegraphics[width=2.3cm]{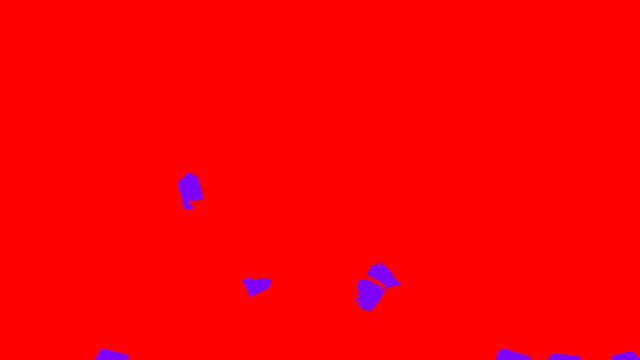}
}
\caption{Comparison of motion feature with and without signature.}
\label{fig:comparison}
\end{figure}

By marking the difference by means of the signature, motion features are more detectable in the presence of missing trajectories, which is demonstrated experimentally in Figure~\ref{fig:comparison}. The subspace clustering method succeeded in segmenting motions with signature (\ref{fig:with_signature}) while it fails in the absence of signature (\ref{fig:without_signature}).

\subsection{Unified optimization framework}
We propose the following unified optimization framework \eqref{pb:sssc-am-1}. In order to make it work on affine subspaces, we add an affine subspace constraint $Z_k^{\top} \tmmathbf{1}=\tmmathbf{1}$ to the original $\mathrm{S^{3}C}$ framework \eqref{eqn:opt_pb}, and formulate the optimization problem to solve as follows:
\begin{eqnarray*}{C}
  \label{pb:sssc-am-1} \min_{\substack{Z_1, Z_2, \ldots, Z_K \\ E_1, E_2, \ldots, E_K \\ Q_1, Q_2, \ldots, Q_K}} 
  \sum^K_{k = 1} (\| Z_k \|_{1, Q_{k}} + \lambda_k \| E_k \|_1) + \beta \| Z
  \|_{2, 1}  \\
  s.t. \quad X_k = X_k Z_k + E_k, \; \tmop{diag} (Z_k) =\tmmathbf{0}, \\
  Z_k^{\top} \tmmathbf{1}=\tmmathbf{1},  \; Q_{k} \in \mathcal{Q} 
\end{eqnarray*}{
where
\begin{eqnarray*}
  Z & = & \left(\begin{array}{cccc}
    \tmop{vec} (Z_1^{\top}) & \tmop{vec} (Z_2^{\top}) & \cdots & \tmop{vec}
    (Z_K^{\top})
  \end{array}\right)^{\top}\\
  & = & \left[\begin{array}{cccc}
    (Z_1)_{11} & (Z_1)_{12} & \cdots & (Z_1)_{\tmop{NN}}\\
    (Z_2)_{11} & (Z_2)_{12} & \cdots & (Z_2)_{\tmop{NN}}\\
    \vdots & \vdots &  & \vdots\\
    (Z_K)_{11} & (Z_K)_{12} & \cdots & (Z_K)_{\tmop{NN}}
  \end{array}\right]_{K \times N^2}
\end{eqnarray*}
and $\| Z_k \|_{1, Q_{k}} = \| Z_k \|_1 + \alpha \| \Theta_{k} \odot Z_k \|_1$.
$\tmmathbf{0}$ and $\tmmathbf{1}$ are two vectors of appropriate dimensions
filled with $0$ and $1$, respectively. $X_{k}$ refers to different feature data points. Here, with $K=2$, $X_{1}$ denotes appearance features and $X_{2}$ denotes motion features. $Z_{k}$ is the coefficient matrix and $E$ a matrix of errors. The constraint $Z_k^{\top} \tmmathbf{1}=\tmmathbf{1}$ generalizes the framework to affine subspaces by writing each data point as an affine combination of a few other points. Since the linear subspace is a special case of the affine subspace, our framework works effectively also on data points lying in linear subspaces.

$\| Z \|_{2, 1}$ is a penalty term which encourages a unified affinity matrix inferred from appearance and motion features since the $\ell_{2,1}$ norm can induce column sparsity of the matrix \cite{Wang2014}. 

\begin{algorithm}[t]
\caption{Unified framework for solving problem \eqref{pb:sssc-am-1} with affine subspace constraint}
\label{algo:SSSC_affine}
\algsetup{linenodelimiter=.}
\begin{algorithmic}[1]
\REQUIRE Data matrix $X$, $\alpha$, $\beta$, $\lambda$, $\rho$, and the number of subspaces.
\STATE \textbf{Initialize:} $\Theta_{0}=\bm{0}$, $\epsilon=10^{-6}$
\WHILE{not converged}
\STATE Given $\Theta_{T}$, initialize $E=\bm{0}$, $C=Z=J=\bm{0}$, $Y^{(1)}=\bm{0}$, $Y^{(2)}=Y^{(3)}=\bm{0}$, $Y^{(4)}=\bm{0}$
\WHILE{not converged}
\STATE Update $J_{t}$, $C_{t}$, $E_{t}$, and $Z_{t}$ according to \eqref{eqn:J} \eqref{eqn:C} \eqref{eqn:E} \eqref{eqn:Z};
\STATE Update $Y^{(1)}_{t}$, $Y^{(2)}_{t}$, $Y^{(3)}_{t}$, $Y^{(4)}_{t}$ according to \eqref{eqn:Y};
\STATE Update $\mu_{t+1}\leftarrow\rho\mu_{t}$;
\STATE Check the condition $\norm{X-XC-E}_{\infty}<\epsilon$; if not converged, then set $t\leftarrow t+1$; if converged, obtain $Z$
\ENDWHILE
\STATE Given $Z$, construct the affinity graph $G$ given by $\mathbf{W}=\lvert\mathbf{Z}\rvert+\lvert\mathbf{Z}\rvert^{\mathrm{T}}$.
\STATE Apply the spectral clustering technique to the affinity graph with the input number of subspaces as $k$ and obtain $\Theta_{T+1}$.
\STATE Check the convergence condition $\norm{\Theta_{T+1}-\Theta_{T}}_{\infty}<1$; if not converged, $T\leftarrow T+1$
\ENDWHILE
\ENSURE Coefficient matrix $Z$ and the label $l$ indicating subspaces where data points lie in
\end{algorithmic}
\end{algorithm}

To solve the optimization problem \eqref{pb:sssc-am-1}, we notice that it is equivalent to the following
problem \eqref{pb:sssc-am-2} which can be solved more efficiently.
\begin{eqnarray*}{C}
 \label{pb:sssc-am-2} \min_{\substack{Z_1, Z_2, \ldots, Z_K \\ E_1, E_2, \ldots, E_K \\ Q_1, Q_2, \ldots, Q_K}}
  \sum^K_{k = 1} (\| J_k \|_{1, Q_k} + \lambda_k \| E_k \|_1) + \beta \| Z
  \|_{2, 1}   \\
  s.t. \quad X_k = X_k C_k + E_k, \; J_k =  Z_k, \\
  C_k = J_k - \tmop{diag} (J_k),\;  C_k^{\top} \tmmathbf{1}=\tmmathbf{1}, \; Q_{k} \in \mathcal{Q}  
\end{eqnarray*}

We can solve this problem using the Alternating Direction Method
of Multipliers (ADMM) method \cite{Boyd2010}, which is a powerful yet simple algorithm well suited to problems arising in applied statistics and machine learning.
The augmented Lagrangian is given by:
\begin{eqnarray*}{l}
  \mathcal{L} (J_k, C_k, E_k, Z, Y_k^{(1)}, Y_k^{(2)}, Y_k^{(3)}, Y_k^{(4)} ;
  k = 1, \ldots, K) \\
  = \sum_{k = 1}^K \| J_k \|_{1, Q_k} + \lambda_k \| E_k \|_1 + \langle
  Y_k^{(1)}, X_k - X_k C_k - E_k \rangle \\
  + \langle Y_k^{(2)}, C_k - J_k + \tmop{diag} (J_k) \rangle + \langle
  Y_k^{(3)}, J_k - Z_k \rangle \\
  + \langle Y_k^{(4)}, C_k^{\top}
  \tmmathbf{1}-\tmmathbf{1} \rangle + \frac{\mu}{2} (\| X_k - X_k C_k - E_k
  \|_F^2 \\
  + \| C_k - J_k + \tmop{diag} (J_k) \|_F^2 + \| J_k - Z_k \|_F^2 \\
  + \|
  C_k^{\top} \tmmathbf{1}-\tmmathbf{1} \|_F^2) + \beta \| Z \|_{2, 1},
\end{eqnarray*}
where $Y_k^{(1)}, Y_k^{(2)}, Y_k^{(3)}, Y_k^{(4)}$ are Lagrange multipliers.
ADMM allows us to update variables alternatively. In addition, since $J_k,
C_k, E_k, Y_k^{(1)}, Y_k^{(2)}, Y_k^{(3)}, Y_k^{(4)} \tmop{with} \; k = 1,
\ldots, K$ are separable in the optimization problem, we can update them
separately given $k$ (using OpenMP for multi-cores processors).\\
\textbf{Update $J_k$}. We update $J_k$ by solving the following problem:
\begin{eqnarray*}{lll}
  J_{k, t + 1} &=& \underset{J_k}{\arg \min} \frac{1}{\mu_t} \| J_k \|_{1, Q_k} +
  \frac{1}{2} \| J_k - \tmop{diag} (J_k) - U_{k, t} \|_F^2 \\
  &+& \frac{1}{2} \|J_k - Z_{k, t} + Y_{k, t}^{(3)} / \mu_t \|^2_F  ,
\end{eqnarray*}
where $\| J_k \|_{1, Q_k} = \| J_k \|_1 + \alpha \| \Theta_k \odot J_k \|_1$ and
$U_{k, t} = C_{k, t} + \frac{1}{\mu_{t}} Y_{k, t}^{(2)}$.\\
The closed-form solution for $J_k$ is given by
\begin{eqnarray*}{l}
\label{eqn:J}
  J_{k, t + 1}^{i, j} = \left\{
  	\begin{array}{ll}
  \frac{1}{2} S_{\frac{1}{\mu_t} (1 + \alpha \Theta_k^{i, j})} \left( U_t^{i, j} + Z_{k, t}^{i, j} - \frac{Y_{k, t}^{(3), i, j}}{\mu_t} \right) & i\neq j, \\
  S_{\frac{1}{\mu_t} (1 + \alpha \Theta_k^{i, j})} \left( Z_{k, t}^{i, j} - \frac{Y_{k, t}^{(3), i, j}}{\mu_t} \right) & i = j.
\end{array} \right. 
\end{eqnarray*}
where $U_{k, t} = C_{k, t} + \frac{1}{\mu} Y_{k, t}^{(2)}$. $S_{\tau}(\cdot)$ is the shrinkage thresholding operator acting on every element of a matrix. It is defined as: $S_{\tau} (\nu) = (| \nu | - \tau)_+ \tmop{sgn} (\nu)$ with the operator
$(\nu)_+$ returning $\nu$ if it is non-negative and returning zero otherwise.
\\
\textbf{Update $C_k$}. We update $C_k$ by solving the following problem:
\begin{eqnarray*}{lll}
  C_{k, t + 1} &=& \underset{C_k}{\arg \min} \langle Y_{k, t}^{(1)}, X_k - X_k
  C_k - E_{k, t} \rangle \\
  &+& \langle Y_k^{(2)}, C_k - J_{k, t + 1} + \tmop{diag} (J_{k, t + 1}) \rangle\\
  &+& \langle Y_{k, t}^{(4)}, C_k^{\top}\tmmathbf{1}-\tmmathbf{1} \rangle 
  + \frac{\mu_t}{2} (\| C_k^{\top} \tmmathbf{1}-\tmmathbf{1} \|_F^2  \\
  &+& \| C_k - J_{k, t + 1} + \tmop{diag} (J_{k, t + 1}) \|_F^2  \\
  &+&\| X_k - X_k C_k - E_{k, t} \|_F^2)  
\end{eqnarray*}
The solution  is given by
\begin{eqnarray*}{lll}
\label{eqn:C}
  C_{k, t + 1}& =& (X_k^{\top} X_k + \Iota +\tmmathbf{1}\tmmathbf{1}^{\top})^{-1} \bigg[ X_k^{\top} \left( X_k - E_{k, t} + \frac{1}{\mu_t} Y_{k, t}^{(1)} \right) \\
  &+& J_{k, t + 1} - \tmop{diag} (J_{k, t + 1}) - \frac{1}{\mu_t} Y_{k,  t}^{(2)}  \\
  &-& \frac{1}{\mu_t} (\tmmathbf{1}Y_{k, t}^{(4) \top})
  +\tmmathbf{1}\tmmathbf{1}^{\top} \bigg] 
\end{eqnarray*}
\textbf{Update $E_k$}. We update $E_k$ by solving the following problem:
\begin{eqnarray}
  E_{k, t + 1} = \underset{E_k}{\arg \min} \frac{\lambda_k}{\mu_t} \| E_k \|_1
  + \frac{1}{2} \| E_k - V_{k, t} \|^2_F, &  & 
\end{eqnarray}
where $V_{k, t} = X_k - X_k C_{k, t + 1}$+$\frac{1}{\mu_t} Y_{k, t}^{(1)}$.\\
Then the solution is given by
\begin{eqnarray}
\label{eqn:E}
  E_{k, t + 1} = \underset{}{} & S_{\frac{\lambda}{\mu_t}} (V_{k, t}) & 
\end{eqnarray}
where $V_{k, t} = X_k - X_k C_{k, t + 1}$+$\frac{1}{\mu_t} Y_{k, t}^{(1)}$.\\
\textbf{Update $Z$}. We solve $Z$ by solving the following problem:
\begin{eqnarray}
  Z_{t + 1} = & \underset{Z}{\arg \min} \frac{\beta}{\mu_t} \| Z \|_{2, 1} +
  \frac{1}{2} \| Z - P_t \|^2_F, & 
\end{eqnarray}
where
\begin{eqnarray*}
  P_t & = & \left[\begin{array}{cccc}
    \tmop{vec} (P_1^{\top}) & \tmop{vec} (P_2^{\top}) & \cdots & \tmop{vec}
    (P_K^{\top})
  \end{array}\right]^{\top}\\
  & = & \left[\begin{array}{cccc}
    (P_1)_{11} & (P_1)_{12} & \cdots & (P_1)_{\tmop{NN}}\\
    (P_2)_{11} & (P_2)_{12} & \cdots & (P_2)_{\tmop{NN}}\\
    \vdots & \vdots &  & \vdots\\
    (P_K)_{11} & (P_K)_{12} & \cdots & (P_K)_{\tmop{NN}}
  \end{array}\right]_{K \times N^2}
\end{eqnarray*}
with $P_{k, t} = J_{k, t + 1} + \frac{Y_{k, t}^{(3)}}{\mu_t}$\\
Then the problem can be solved via Lemma 4.1 of \cite{Liu2010} as
follows:
\begin{eqnarray*}{l}
\label{eqn:Z}
  [Z_{t + 1}]_{:, i} = \left\{
  	\begin{array}{cl}
 \frac{\| [P_t]_{:, i} \|_2 - \frac{\beta}{\mu_t}}{\|
  [P_t]_{:, i} \|_2} [P_t]_{:, i} & \tmop{if} \| [P_t]_{:, i} \|_2 >
  \frac{\beta}{\mu_t}, \\
  0 & \tmop{otherwise}.
\end{array} \right. 
\end{eqnarray*}
where $[Z_{t + 1}]_{:, i}$ and $[P_t]_{:, i}$ refer to the $i^{\tmop{th}}$
column of respective matrix. $Z_{k, t + 1}$ can then be obtained from $Z_{t +
1}$.\\
\textbf{Update $Y_k^{(1)}, Y_k^{(2)}, Y_k^{(3)}, Y_k^{(4)}$}. We update the Lagrange
multipliers by a simple gradient ascent step.
\begin{eqnarray*}{lcl}
\label{eqn:Y}
  Y_{k, t + 1}^{(1)} &= &Y_{k, t}^{(1)} + \mu_t (X_k - X_k C_{k, t + 1} - E_{k,
  t + 1})\\
  Y_{k, t + 1}^{(2)}& = &Y_{k, t}^{(2)} + \mu_t (C_{k, t + 1} - J_{k, t + 1} +
  \tmop{diag} (J_{k, t + 1}))  \\
  Y_{k, t + 1}^{(3)}& =& Y_{k, t}^{(3)} + \mu_t (J_{k, t + 1} - Z_{k, t + 1})  \\
  Y_{k, t + 1}^{(4)} &= &Y_{k, t}^{(4)} + \mu_t (C_{k, t + 1}^{\top}
  \tmmathbf{1}-\tmmathbf{1}) 
\end{eqnarray*}
Finally, we update $\mu$ by $\mu_{t + 1} \leftarrow \rho \mu_t$.
The unified framework is summarized in Algorithm \ref{algo:SSSC_affine}. For the details of the derivation, we refer the readers to \cite{Boyd2010}.

\subsection{Co-segmentation algorithm}
Finally, in order to combine appearance and motion features, we construct the affinity matrix $S$ given by 
\begin{eqnarray}{l}
\label{eqn:S}
(S)_{ij}=\frac{1}{2}\left(\sqrt{\sum_{k=1}^{K}(Z_{k})_{ij}^{2}}+\sqrt{\sum_{k=1}^{K}(Z_{k})_{ji}^{2}}\right)  
\end{eqnarray}
Then, we apply the spectral clustering technique \cite{Luxburg2007} to the affinity matrix $S$ to obtain segmentation results in the form of an array of label $l$ indicating corresponding subspace of each data point. For clarity, we summarize the $\mathrm{S^{3}C-AM}$ algorithm in Algorithm \ref{algo:SSSC_AM}. The overall processing is summarized in Figure \ref{fig:flow}.
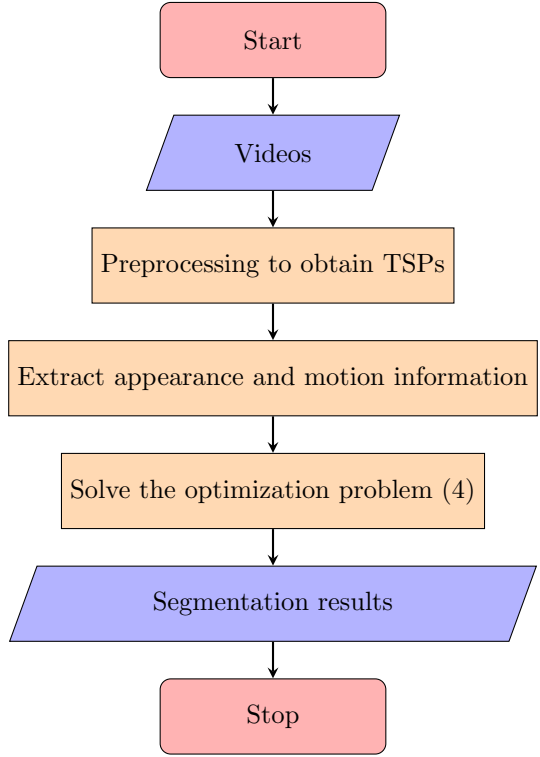
\begin{figure}[t]
\centering
\begin{tikzpicture}[node distance=1.5cm]
\node (start) [startstop] {Start};
\node (in1) [io, below of=start] {Videos};
\node (pro1) [process, below of=in1] {Preprocessing to obtain TSPs};
\node (pro2) [process, below of=pro1] {Extract appearance and motion information};
\node (pro3) [process, below of=pro2] {Solve the optimization problem \eqref{pb:sssc-am-1}};
\node (out1) [io, below of=pro3] {Segmentation results};
\node (stop) [startstop, below of=out1] {Stop};
\draw [arrow] (start) -- (in1);
\draw [arrow] (in1) -- (pro1);
\draw [arrow] (pro1) -- (pro2);
\draw [arrow] (pro2) -- (pro3);
\draw [arrow] (pro3) -- (out1);
\draw [arrow] (out1) -- (stop);
\end{tikzpicture}
\caption{Flow chart of the overall processing}
\label{fig:flow}
\end{figure}

\begin{algorithm}[t]
\caption{$\mathrm{S^{3}C-AM}$}
\label{algo:SSSC_AM}
\algsetup{linenodelimiter=.}
\begin{algorithmic}[1]
\REQUIRE Feature matrices $X_{k},\: k=1,\ldots, K$, $\alpha$, $\beta$, $\lambda$, $\rho$, and the number of subspaces.
\STATE Obtain $Z_{k},\: k=1,\ldots, K$ via Algorithm \ref{algo:SSSC_affine};
\STATE Form the affinity matrix $S$ according to \eqref{eqn:S};
\STATE Apply the spectral clustering technique to the affinity matrix $S$ with the input number of subspaces;
\ENSURE The label $l$ indicating subspaces where TSPs lie in
\end{algorithmic}
\end{algorithm}

\section{Experiments}
\label{sec:experiments}
In this section, we conduct experiments on the dataset Multi-Object Video Co-Segmentation (MOViCS)\footnote{The MOViCS dataset can be found at \url{http://www.d2.mpi-inf.mpg.de/datasets}} proposed by~\cite{Chiu2013} to evaluate the effectiveness of the $\mathrm{S^{3}C-AM}$ framework. This dataset is composed of four sets of real videos, each of which contains different combinations of animal objects: a) chicken-turtle; b) giraffe-elephant; c) zebra-lion; d) tiger. The objects in the sets of videos present various appearance and motion patterns, making the co-segmentation task more challenging. Some animals may move inseparably together such as in the \emph{chicken on turtle} video. Natural coloring of some other animals, including elephant or tiger, provides camouflage so that they blend in perfectly with the surroundings. Several other difficulties include different lighting conditions, and motion blur. Despite those challenges, our experimental results show that our framework achieves the highest overall performance, and outperforms baseline algorithms in two sets of videos and yields comparable results on two remaining video sets. 

\subsection{Parameter setting}
In $\mathrm{S^{3}C-AM}$, the parameters $\lambda_k$ and $\alpha$ need to be set properly as in~\cite{Elhamifar}. 
Parameters $\rho$ and $\beta$ are set to be $1.2$ and $1\times 10^{-5}$, respectively, which work well in practice.

\subsection{Evaluation metric}
In order to evaluate the effectiveness of our framework, we use the following metric:
\begin{equation}
  \tmop{Score} = \frac{1}{C} \sum_j \max_i \frac{S_i \cap G_j}{S_i \cup G_j},
  \label{eq:metric}
\end{equation}
where $G_j$ and $S_i$ refer to sets of segments belonging to $j^{\tmop{th}}$
class obtained by our co-segmentation algorithm, and $i^{\tmop{th}}$ class in ground truth, respectively. $C$ is the
number of classes in ground truth. By computing the average accuracy that a set of segments matches best each object in video frames, the \tmtextit{intersection-over-union
metric} \eqref{eq:metric} is able to reflect the overall performance of a video
co-segmentation method.

\subsection{Results}
We conduct experiments on all video sets in the dataset and compare our framework to other state-of-the-art methods~\cite{Chiu2013} and~\cite{Wang2014}. For the purpose of clarity, we denote the \emph{multi-class video co-segmentation} framework of~\cite{Chiu2013} by VCS and the \emph{video object co-segmentation} framework of~\cite{Wang2014} by VOCS. We refer to our framework as $\text{S\tmrsup{3}C-AM}$. For both VCS and VOCS, we use their publicly available accuracies measured also by the metric \eqref{eq:metric} on the MOViCS dataset.

\begin{figure}[t]
\begin{center}
\includegraphics[width=0.95\linewidth]{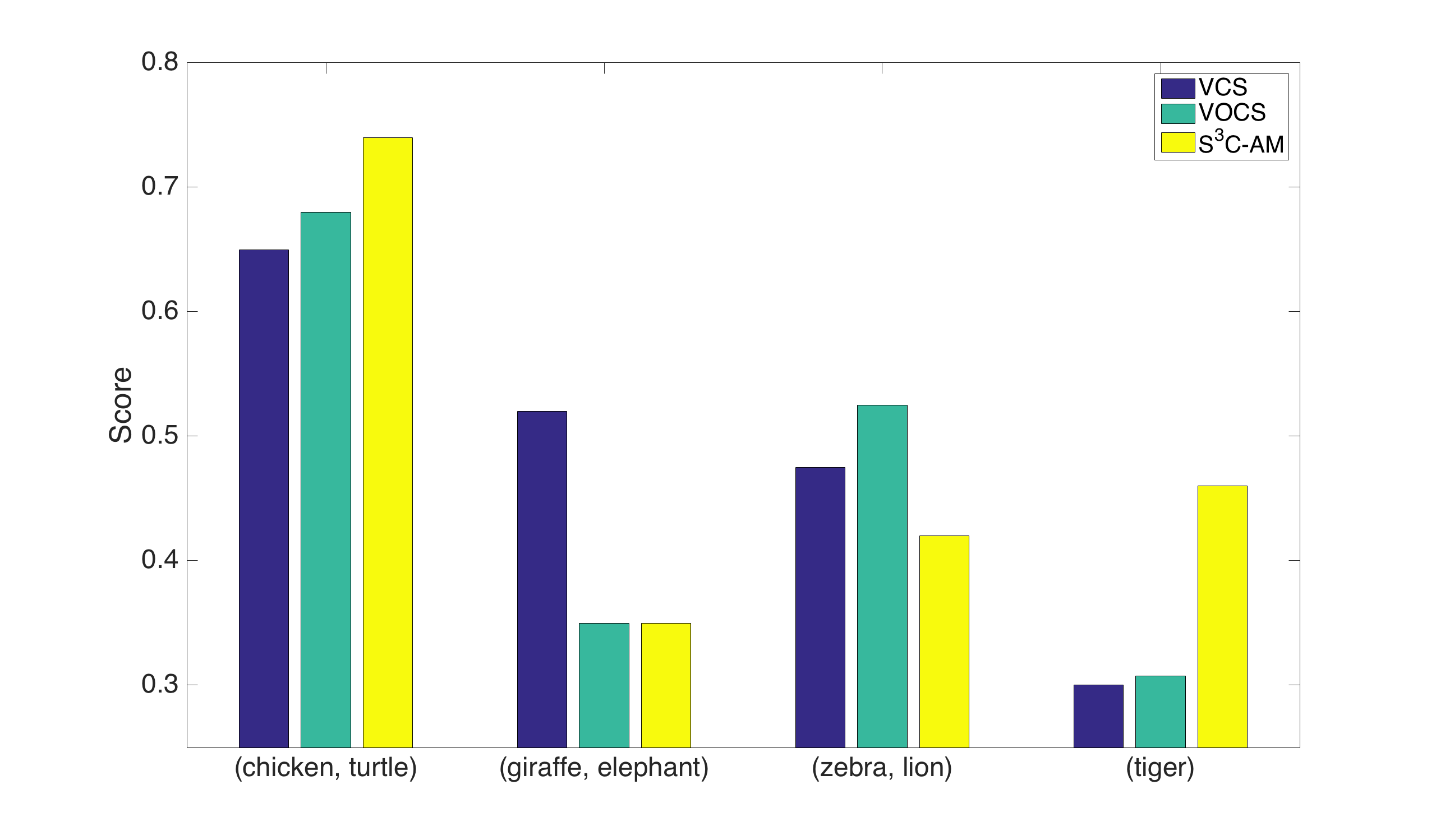}
\caption{Comparison of co-segmentation accuracies between our method ($\text{S\tmrsup{3}C-AM}$),  multi-class video co-segmentation (VCS) and video object co-segmentation (VOCS) on MOViCS dataset.}
\label{fig:accuracyComparison}
\end{center}
\end{figure}

Figure \ref{fig:accuracyComparison} shows the comparison between our framework and the baselines. Achieving an overall performance of 49.25\% by our method, we outperform VCS (48.75\%) and VOCS (46.56\%). For two out of four video sets, i.e., chicken-turtle, and tiger videos, our framework produces better results, particularly for the tiger video set where undistinguishable appearance of the tiger with respect to its surroundings is essentially present. For the giraffe-elephant video set, VCS outperforms both $\text{S\tmrsup{3}C-AM}$ and VOCS while for the zebra-lion video set our method produces lower accuracy compared to two baselines. We will give full discussion in the later subsection.

In addition, we compare computational cost between VCS, VOCS, and our framework. We evaluate our framework under the following computer configuration: 2.7 GHz Intel Core i5, with 8 GB RAM and OS platform, as shown in Table \ref{tb:computationalCost}. Our preprocessing performs superpixel over-segmentation which reduces greatly the number of data points involved in clustering. While~\cite{Wang2014} adopts also TSP-based preprocessing, our method is able to achieve reliable co-segmentation results using less TSPs, leading to faster preprocessing. For three out of four video sets our clustering routine runs much faster than VCS. Due to our iterative clustering algorithm intended for more coherent segmentation, it is expected that our method needs more time in clustering part than VOCS.

\begin{table}[htbp]
\centering
\resizebox{\linewidth}{!}{
  \begin{tabular}{*7c}
  \toprule
    & \multicolumn{3}{c}{Preprocessing} &  \multicolumn{3}{c}{Clustering} \\
    \midrule
    {} & VCS & VOCS & S\tmrsup{3}C-AM & VCS & VOCS & S\tmrsup{3}C-AM\\
    (chicken, turtle) & 1h 40m & 1h 10m & 31m  & 1h 12m & 1h 04m 13s & 1h 18m\\
    (giraffe, elephant) & 1h 33m & 54m & 22m & 1h 22m & 13m 15s & 35m\\
    (zebra, lion) & 3h 19m & 2h 14m & 56m & 3h 13m & 1h 23m 20s & 2h 11m\\
    (tiger) & 1h 11m & 48m & 21m & 59m & 30m 18s & 46m \\
    \bottomrule
  \end{tabular}}
  \caption{\label{tb:computationalCost} Comparison of computation cost between VCS, VOCS, and our method $\text{S\tmrsup{3}C-AM}$. The runtime data of \cite{Chiu2013} and \cite{Wang2014} are obtained from their supplementary material.}
\end{table}

Table \ref{tb:exampleResults} shows visualization of our results. Every block corresponds to frame-result pairs of a video set. First column in each block is a frame of the video, whereas second column shows the results of our method

\begin{table}[!htbp]
\centering
\begin{tabular}{c c | c c}
\toprule
Frame & $\text{S\tmrsup{3}C-AM}$ & Frame & $\text{S\tmrsup{3}C-AM}$
\\
\midrule
\multicolumn{2}{c}{(Chicken, turtle)} & \multicolumn{2}{c}{(Giraffe, elephant)}
\\
\includegraphics[width=0.2\linewidth]{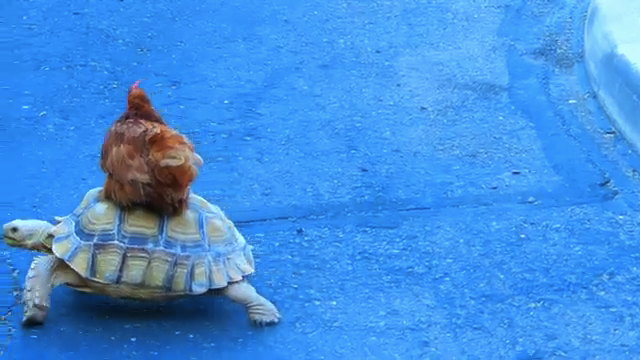} 
& \includegraphics[width=0.2\linewidth]{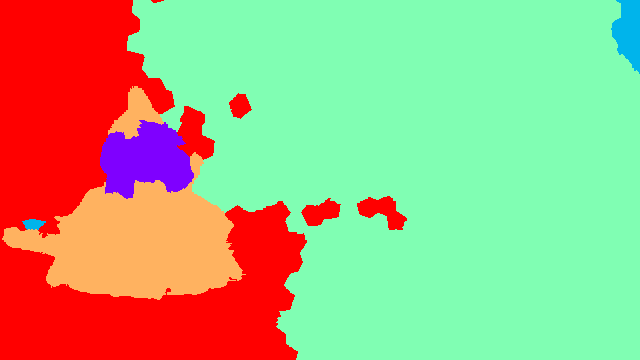}
& \includegraphics[width=0.2\linewidth]{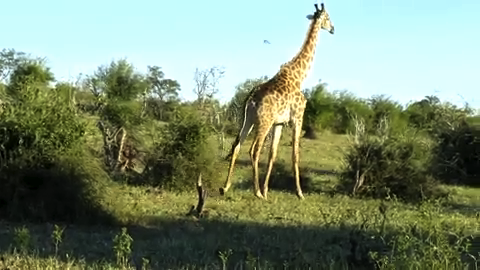}
& \includegraphics[width=0.2\linewidth]{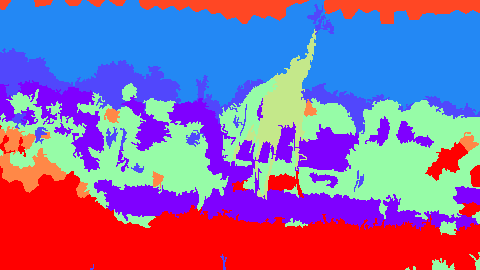}
\\
\includegraphics[width=0.2\linewidth]{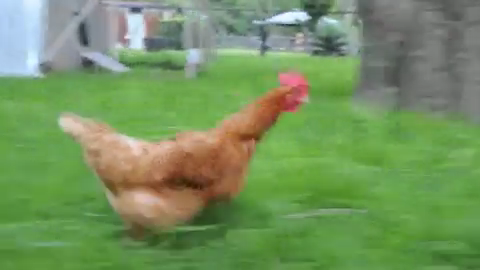}
& \includegraphics[width=0.2\linewidth]{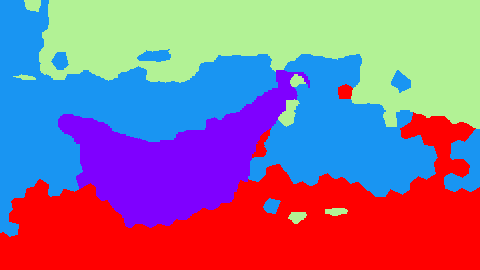} 
& \includegraphics[width=0.2\linewidth]{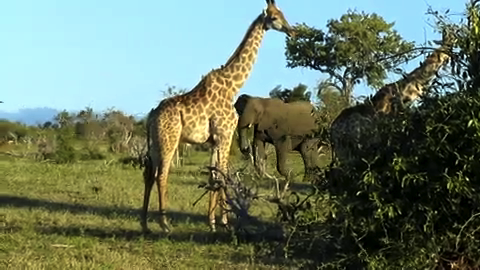} 
& \includegraphics[width=0.2\linewidth]{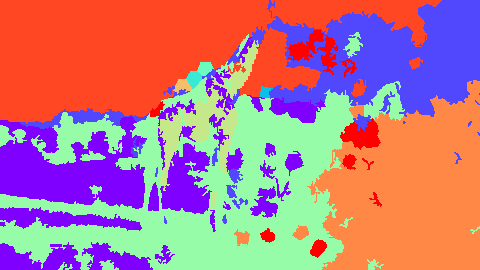} 
\\ 
\midrule
\multicolumn{2}{c}{(Lion, zebra) } & \multicolumn{2}{c}{(Tiger)}
\\
\includegraphics[width=0.2\linewidth]{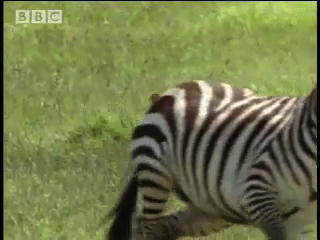}
& \includegraphics[width=0.2\linewidth]{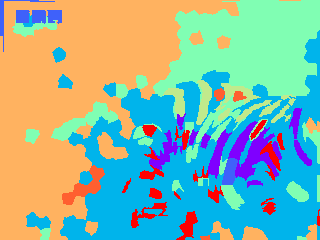} 
& \includegraphics[width=0.2\linewidth]{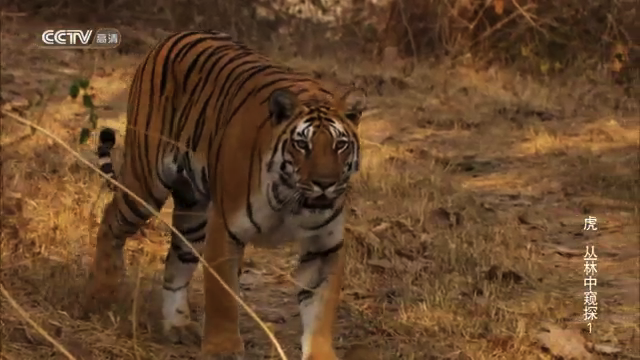} 
& \includegraphics[width=0.2\linewidth]{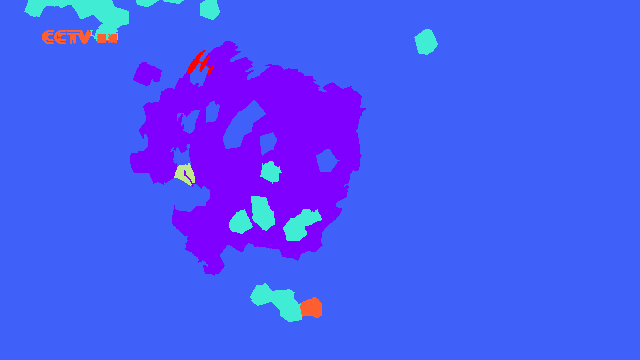} 
\\
\includegraphics[width=0.2\linewidth]{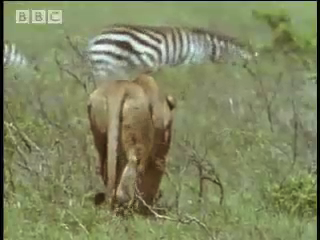}
& \includegraphics[width=0.2\linewidth]{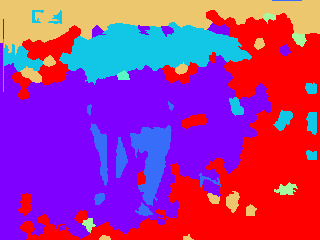} 
& \includegraphics[width=0.2\linewidth]{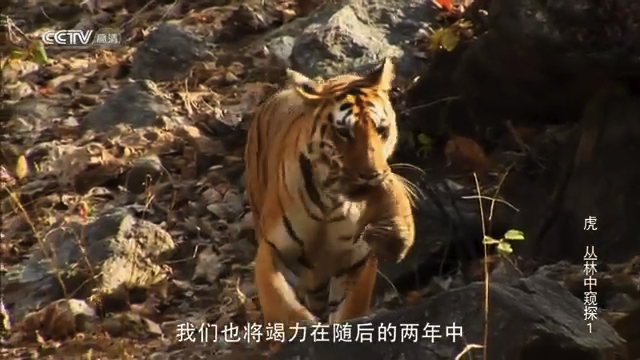}
& \includegraphics[width=0.2\linewidth]{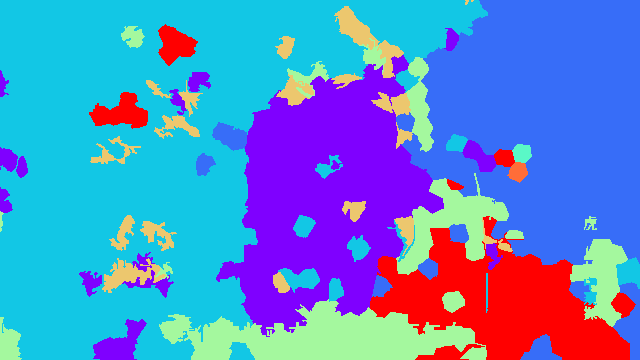}
\\
\includegraphics[width=0.2\linewidth]{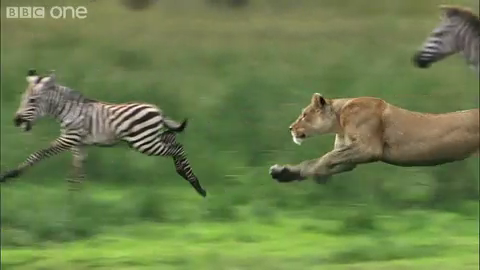}
& \includegraphics[width=0.2\linewidth]{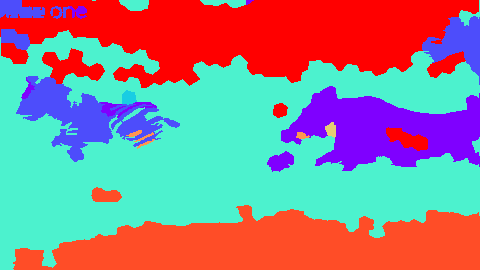}
& \includegraphics[width=0.2\linewidth]{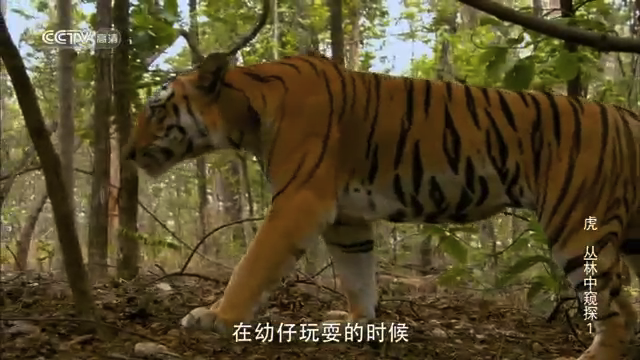}
& \includegraphics[width=0.2\linewidth]{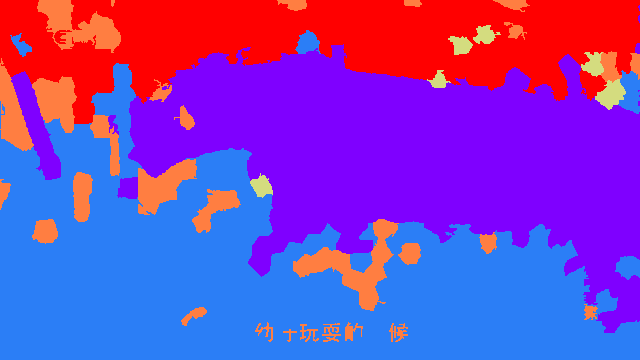}
\\
\includegraphics[width=0.2\linewidth]{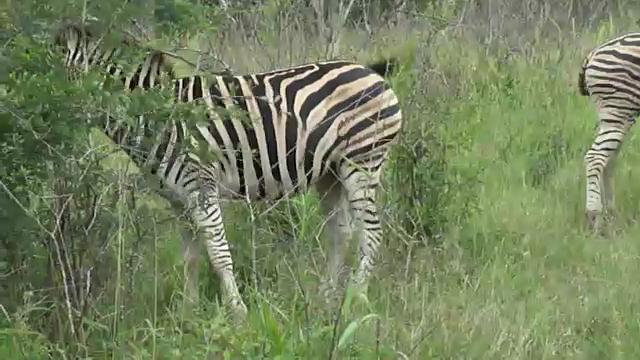}
& \includegraphics[width=0.2\linewidth]{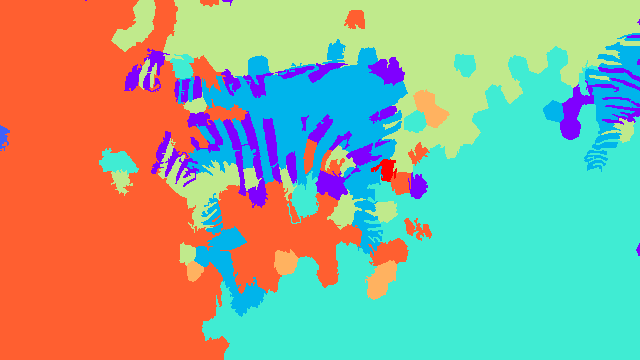} 
& &
\end{tabular}
\caption{Example of results of co-segmentation for all video sets in MOViCS dataset.}
\label{tb:exampleResults}
\end{table}

\subsection{Discussion}
\subsubsection{Appearance and motion features}
Using both appearance and motion information enables our method to better segment objects. For illustration, we compare segmentation of \emph(chicken on turtle) video only with the motion feature (\ref{fig:motion only}) to that solely with the appearance features (\ref{fig:appearance only}). It can be noticed that, with only appearance features, objects with similar colors in the background may be segmented to the same class of the foreground. On the other hand, segmentation without appearance information may result in incomplete segmentation. The combination of appearance and motion features gives a satisfactory segmentation (\ref{fig:clustering_result}) even in the presence of noise and blur found in the original videos.

\begin{figure}[t]
	\centering
	\subfloat[Original frame]{
		\label{fig:original_frame_2}
		\includegraphics[width=3.5cm]{original_image.png}
	}
	\hspace{0.05\linewidth}
	\subfloat[Segmentation result]{
		\label{fig:clustering_result}
		\includegraphics[width=3.5cm]{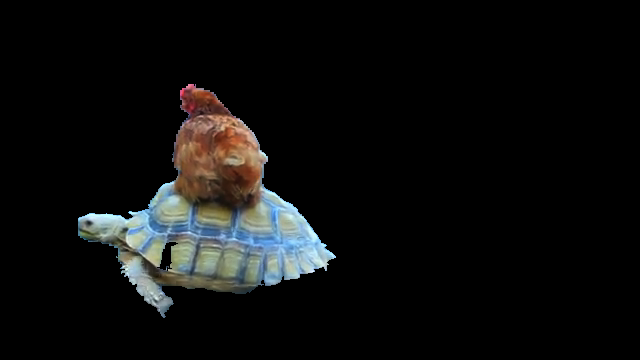}
	}
	\hspace{0.05\linewidth}
	\subfloat[motion only]{
		\label{fig:motion only}
		\includegraphics[width=3.5cm]{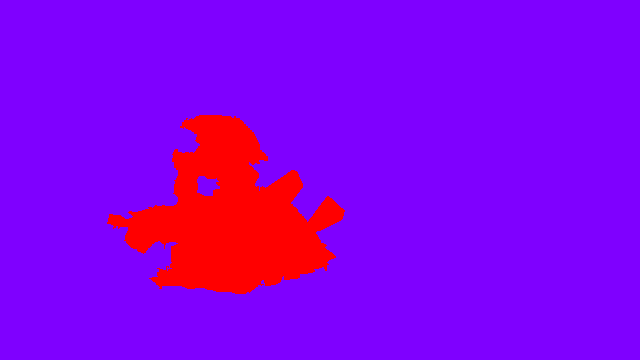}
	}
	\hspace{0.05\linewidth}
	\subfloat[appearance only]{
		\label{fig:appearance only}
		\includegraphics[width=3.5cm]{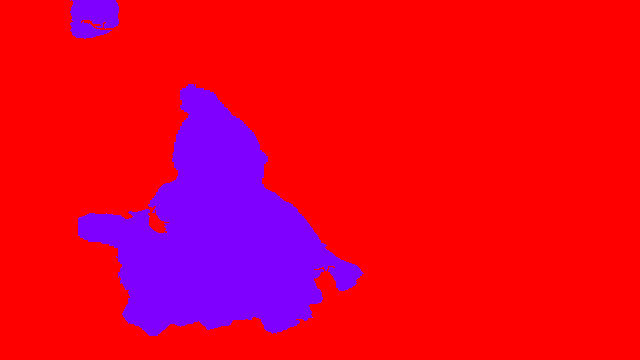}
	}
	\caption{Segmentation results on the \emph{chicken on turtle} video sequences}
	\label{fig:result}
\end{figure}

\subsubsection{Affine subspace constraint}
As stated before, we added an affine subspace constraint $Z_k^{\top} \tmmathbf{1}=\tmmathbf{1}$ to the original $\mathrm{S^{3}C}$ framework \eqref{eqn:opt_pb}, which we believe makes ours yield better results on affine subspaces. In order to highlight the role of the affine subspace constraint with respect to the motion feature, we compare our framework with $\mathrm{S^{3}C}$ on the Hopkins 155 motion segmentation dataset~\cite{Tron2007A-benchmark-for} that allows us to analyze the ability of both frameworks to segment video sequences into multiple spatio-temporal regions corresponding to the different motions in the scene without interference of the appearance feature. In the dataset, feature points are already tracked so that no preprocessing of temporal superpixels is needed. Since our framework is compatible with a single feature by removing the penalty term $\| Z \|_{2, 1}$, we set $\beta$ to be $0$ to ensure an equivalent parameter setting. For both frameworks, we feed motion features without the extra-dimensional signature. 

The evaluation metric of clustering is given by
\begin{equation}
\tmop{error} (l, \hat{l}) = 1 - \max_{\pi} \frac{1}{N} \sum_{i = 1}^N 1_{\{
	\pi (\hat{l}) = l \}}, \label{eq:error}
\end{equation}
where $l, \hat{l} \in \{ 1, \ldots, M \}^N$ are original and estimated labels of tracked points in the $M$ subspaces. As the labels assigned to the points by subspace clustering algorithms may be a permutation of the original ones, we find the most coherent labels by maximizing the second term with respect to all permutations $\pi : \: \{ 1, \ldots, M \}^N \rightarrow \{ 1, \ldots, M \}^N$.

\begin{table}[htbp]
	\centering
		\begin{tabular}{*5c}
			\toprule
			No. motions & \multicolumn{2}{c}{2} &  \multicolumn{2}{c}{3} \\
			Error(\%) & Ave. & Med. & Ave. & Med.\\
			\midrule
			S\tmrsup{3}C~\cite{Li2015a} & 1.94 & \textbf{0} & 4.92 & 0.89 \\
			S\tmrsup{3}C-AM & \textbf{1.53} & \textbf{0} & \textbf{4.40} & \textbf{0.56} \\
			\bottomrule
		\end{tabular}
	\caption{\label{tb:affineConstraint} Comparison of motion segmentation errors on Hopkins 155 Database between S\tmrsup{3}C (without the affine subspace constraint) and S\tmrsup{3}C-AM (with the affine subspace constraint). The best results are in bold font.}
\end{table}

Experimental results of the comparison are presented in Table \ref{tb:affineConstraint}. It is shown that with the additional affine subspace constraint, our framework is able to achieve higher accuracy when segmenting rigid motions lying in affine subspaces. The affine subspace constraint, which improves the performance of motion segmentation, as well as the combination of motion and appearance features, ensure the performance of our framework.

\subsubsection{Problems caused by missing trajectories}
Our framework exploits the superpixel over-segmentation. Thus, the performance of our method depends on results of the TSP preprocessing. While TSP method outperforms other state-of-the-art approaches and provides reliable segmentation results, it produces dead TSPs and new TSPs due to occlusion or tracked points moving into or out of frames. Consequently, TSPs can only survive during a limited number of frames. Such phenomenon would greatly reduce the inter-frame temporal extent of TSPs, resulting in many missing trajectories, especially when objects move rapidly or encounter frequent occlusion caused by themselves or others. This is exactly the case in (giraffe, elephant) and (zebra, lion) video sets. Even though we propose a novel and effective method to deal with missing trajectories through an extra-dimensional signature, we lose too many trajectories to have the signature compensate for the loss of information. Since our method emphasizes essentially the coherence of both appearance and motion features, a flaw in motion features would finally result in failure of the unified framework. Therefore, due to lack of sufficient motion features related to (giraffe, elephant) and (lion, zebra) videos, our method fails to fully discover objects in video sets.

\subsubsection{Multi-class segmentation vs. co-segmentation}
As we understand, multi-class segmentation and co-segmentation both refer to the process of assigning a label to every pixel in videos and partitioning them into meaningful groups. They are closely related concepts while they differ in the way to provide additional information in the absence of supervisory information. Co-segmentation seeks to segment an object in video groups where it all appears. As such, the presence of objects shared among videos provides more information and encourages co-segmentation. For example, given a video of a chicken on a turtle and the other video of only a chicken, our framework is able to extract the common object, i.e., the chicken and thus obtain a refined result as shown in Table~\ref{tb:chicken}. On the other hand, multi-class segmentation does not enforce such setting. It usually discovers object segmentation based on a single video where salient objects are present. The underlying assumption of multi-class segmentation and co-segmentation about the structure of videos may coincide. It is also possible for co-segmentation to assume existence of several classes, or only background/foreground binary classes. Our framework is more like multi-class co-segmentation in the sense that it assumes multi-class structure of videos and yields multi-class segmentation results. It is also possible for our framework to yield binary segmentation if videos are less structured as in the case of (chicken, turtle). In general cases, further processing is needed to highlight foreground/background, such as user guidance. 

\begin{table}[t]
\centering
\begin{tabular}{c c c}
\includegraphics[width=2.5cm]{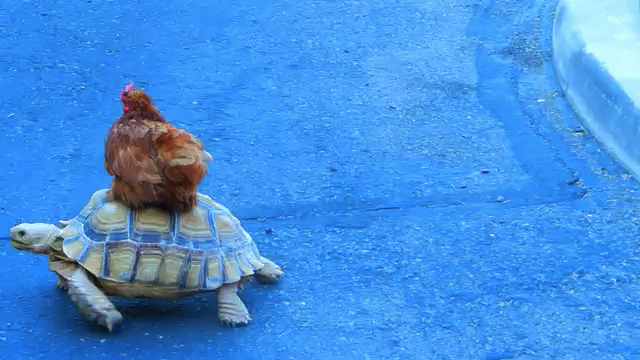} &
\includegraphics[width=2.5cm]{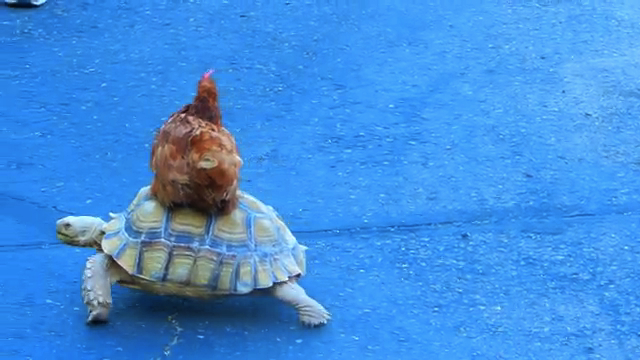} &
\includegraphics[width=2.5cm]{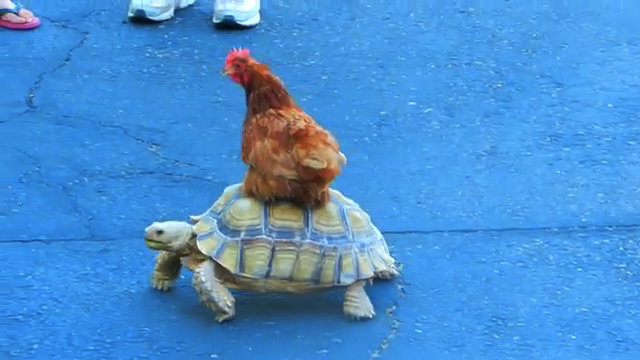}
\\
\includegraphics[width=2.5cm]{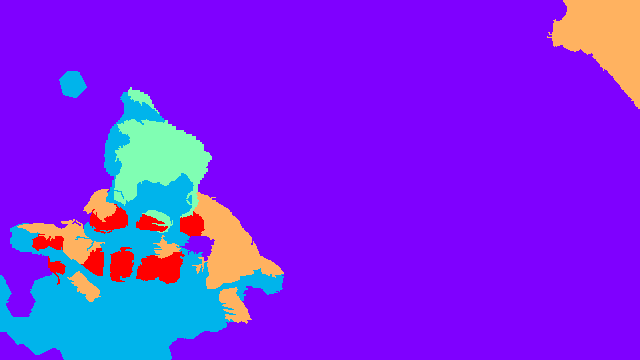} &
\includegraphics[width=2.5cm]{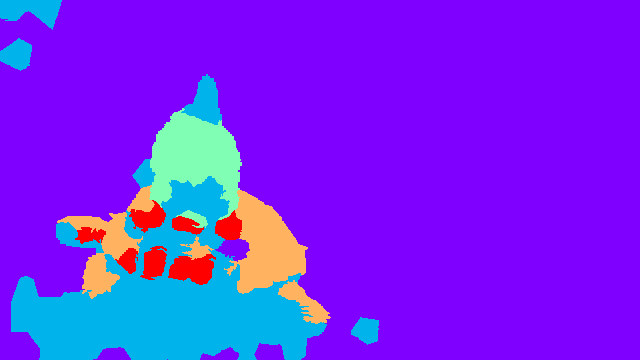} &
\includegraphics[width=2.5cm]{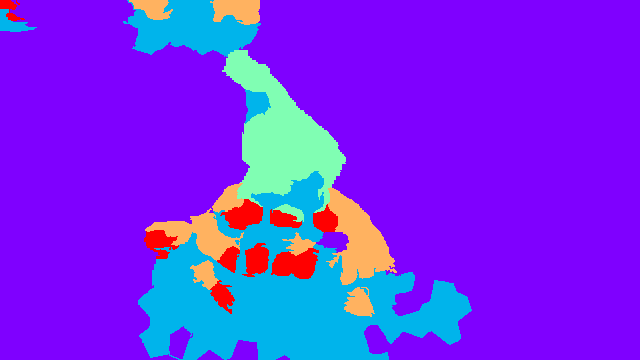}
\\
\includegraphics[width=2.5cm]{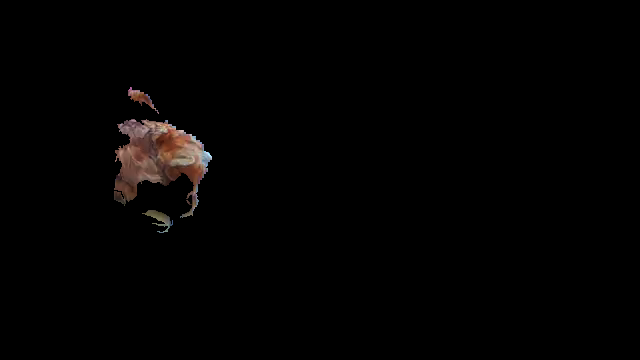} &
\includegraphics[width=2.5cm]{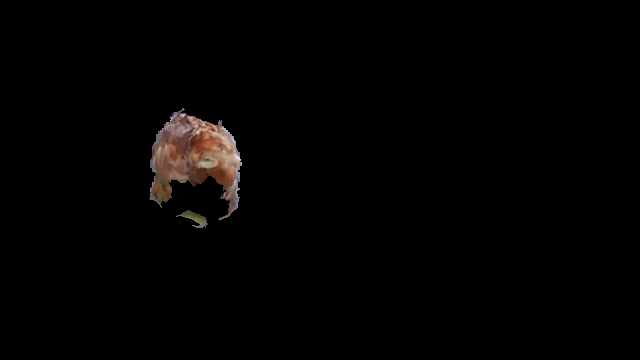} &
\includegraphics[width=2.5cm]{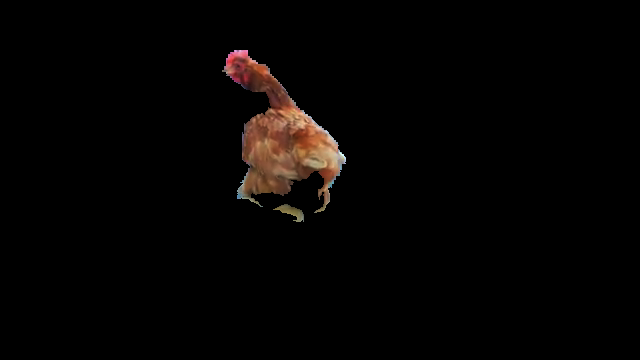}
\\
\includegraphics[width=2.5cm]{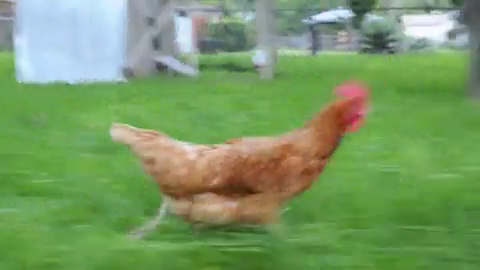} &
\includegraphics[width=2.5cm]{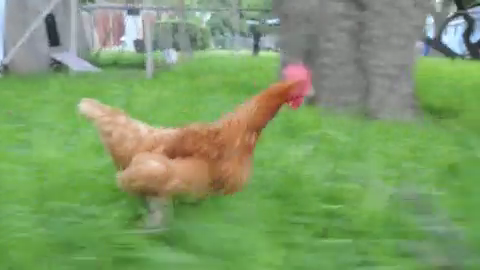} &
\includegraphics[width=2.5cm]{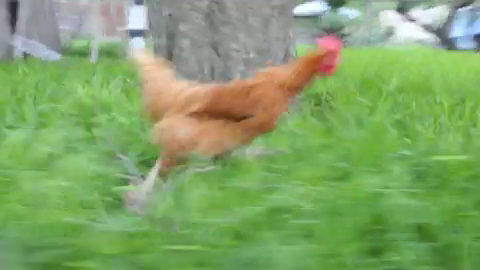} 
\\
\includegraphics[width=2.5cm]{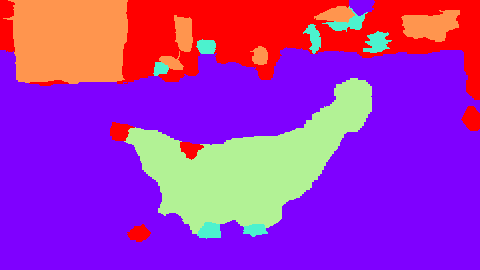} & 
\includegraphics[width=2.5cm]{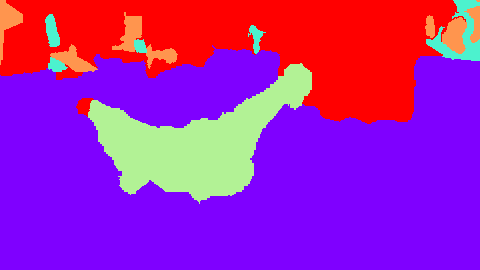} & 
\includegraphics[width=2.5cm]{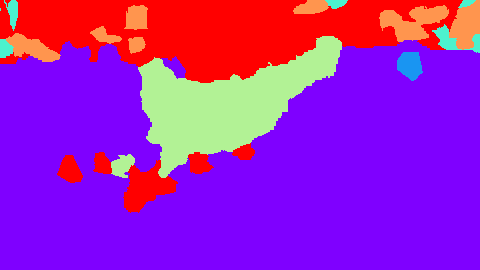}
\\
\includegraphics[width=2.5cm]{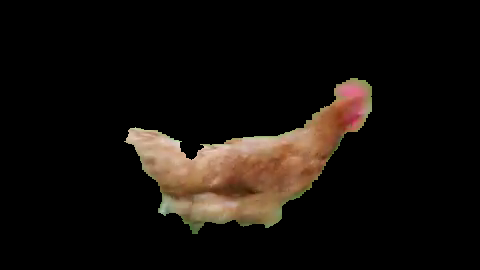} &
\includegraphics[width=2.5cm]{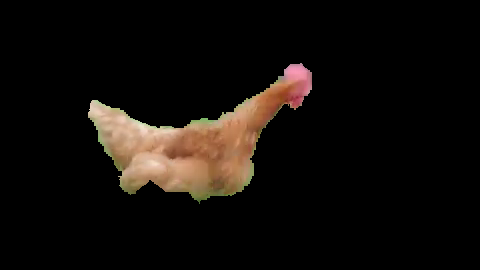} &
\includegraphics[width=2.5cm]{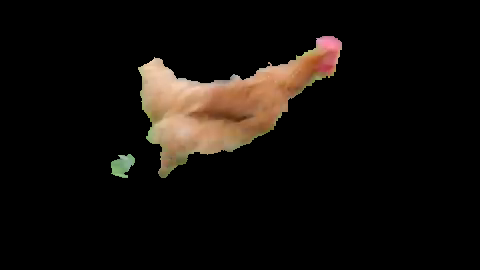}
\end{tabular}
\caption{Results of co-segmentation on videos \emph{chicken on turtle} and \emph{chicken}. Columns 1, 2, and 3 correspond to frames 1, 9, and 17. Rows 1 and 4 correspond to the original video frames, rows 2 and 5 are multi-class segmentation, rows 3 and 6 correspond to the extracted common objects.}
\label{tb:chicken}
\end{table}

\subsubsection{Robustness under the condition of heavy noise}
Finally, we tested the robustness of our framework under the condition of heavy noise. While for most videos there is no concern about resolution and noise, in some circumstances it is inevitable to encounter video quality issues. For example, scenes shot at night under poor lighting conditions are prone to contain heavy noise due to high ISO sensitivity. We added gaussian-distributed additive noise with variance of $0.25$, or $5\%$ amount of salt-and-pepper noise to the original video. It is shown that our framework is effective even with heavy data corruption as illustrated in Figure~\ref{fig:result_noise}.

\begin{figure}[t]
\centering
\subfloat[Original frame corrupted by gaussian-distributed noise]{
	\label{fig:original_frame_noise}
	\includegraphics[width=3.5cm]{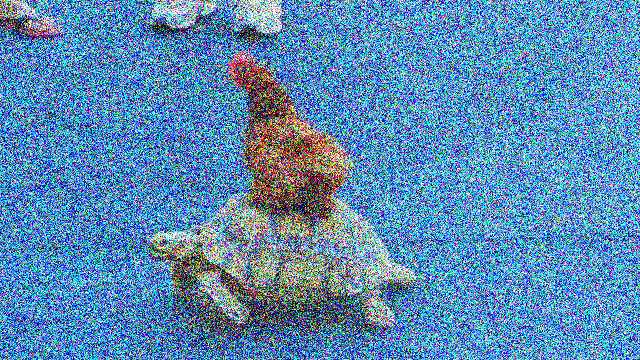}
}
\hspace{0.05\linewidth}
\subfloat[Segmentation result]{
	\label{fig:clustering_result_noise}
	\includegraphics[width=3.5cm]{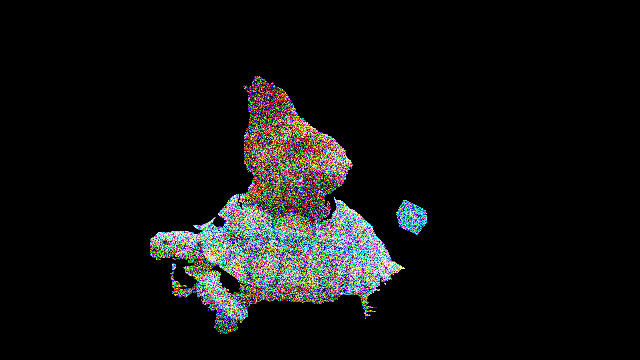}
}
\vspace{0.05\linewidth}
\subfloat[Original frame corrupted by salt-and-pepper noise]{
	\label{fig:original_frame_sp}
	\includegraphics[width=3.5cm]{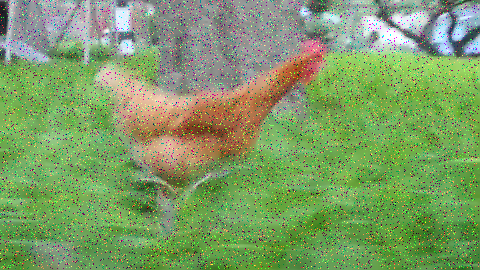}
}
\hspace{0.05\linewidth}
\subfloat[Segmentation result]{
	\label{fig:clustering_result_sp}
	\includegraphics[width=3.5cm]{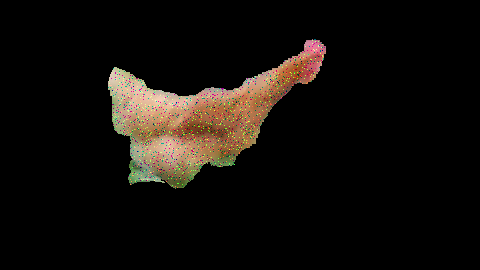}
}
\caption{Segmentation results on corrupted videos}
\label{fig:result_noise}
\end{figure}

\section{Conclusion and perspectives}
\label{sec:conclusion}\label{sec:future}
We described a novel unified framework for video co-segmentation problem. 
By adding an extra-dimensional signature to motion trajectories, we improve the detectability of motion features in the subspace clustering framework. Moreover, we reformulated the $\mathrm{S^3C}$ with the additional affine subspace constraint to successfully segment data points drawn from a union of affine subspaces. We then formulated the unified optimization framework by integrating the newly proposed subspace clustering algorithm and applied it to appearance and motion features. Experiments on the MOViCS benchmark demonstrated the effectiveness of our method and its superiority over baselines. In future work, we would like to extend our approach by generalizing the framework to objects with articulated motion. In addition, we plan also exploit a distributed optimization based on ADMM to improve the running times.
We concisely hint at those extensions as follows:

\subsection{Object with articulated motion}
As shown by the experiments, our approach segments effectively salient objects from videos by jointly making full use of the appearance and motion information. As for the motion information, it is assumed that each object has a rigid-body motion which forms an affine subspace. Though this assumption works well in general, it is difficult to handle objects with articulated motion. To generalize our work, one possible solution is to encode motion information in a different way so as to express subspace structure yet to preserve properties of articulated motion. Another approach consists in exploiting the inherent hierarchical structure of the articulated motion. 

\subsection{Distributed optimization}
Another possible extension of this work consists in taking advantage of the inherent parallel computing structure of ADMM. As demonstrated in the Algorithm~\ref{algo:SSSC_affine}, ADMM takes the form of a \emph{decomposition-coordination} procedure, coordinating solutions to local subproblems to find a solution to a global problem. Therefore, it is natural to develop parallel and distributed optimization algorithm based on ADMM to solve efficiently the large-scale video co-segmentation problem.

A complete source code implementation of {\sc SSSC-AM} in Python for reproducible research is available from the Authors at request.


\newpage
\appendix

\section{Notation}

\begin{supertabular}{p{0.15\columnwidth}p{0.60\columnwidth}}
$A$, $B$ & Arbitrary $m\times n$ matrices \\
$a_{i,j}$ & the $(i,j)$-th element of matrix $A$\\
$a_{i}$ & The $i$-th column of $A$\\
$\mathrm{vec}(A)$ & $mn\times 1$ column vector obtained by stacking the columns of $A$ on top of one another: $\mathrm{vec}(A) = [a_{1,1},\ldots, a_{m,1}, a_{1,2},\ldots, a_{m,2},\ldots,a_{1,n},\ldots, a_{m,n}]^{\top}$ \\
$\norm{A}_{1}$ & $\norm{A}_{1} = \sum_{i=1}^{m}\sum_{j=1}^{n}\vert a_{ij}\vert$ \\
$\norm{A}_{2,1}$ & $\norm{A}_{2,1} = \sum_{j=1}^{n}\norm{a_{j}}_{2} = \sum_{j=1}^{n}\left(\sum_{i=1}^{m}\vert a_{ij}\vert^{2}\right)^{1/2}$ \\
$X$ & $n\times N$ data matrix. Each column refers to a data point. \\
$Z$ & $N\times N$ coefficient matrix\\
$\mathrm{diag}(Z)$ & $\mathrm{diag}(Z)\in \mathbbm{R}^{N}$, vector of the diagonal elements of $Z$ \\
$E$ & $n\times N$ noise matrix \\
$\tmmathbf{0}$ & Vector of appropriate dimensions filled with $0$ \\
$\tmmathbf{1}$ & Vector of appropriate dimensions filled with $1$ \\
$Q$ & $N\times k$ segmentation matrix with $k$ the number of subspaces. $Q=[\mathbf{q_{1}}, \cdots, \mathbf{q_{k}}]$ indicates the membership of each data point to each subspace. $q_{i,j}=1$ if the $i$-th column of $X$ lies in subspace $S_{j}$ and $q_{i,j}=0$ otherwise. \\
$\bm{q}^{(i)}$ & $i$-th row of matrix $Q$ \\
$\mathcal{Q}$ & Space of segmentation matrices: $\mathcal{Q}=\{Q\in \{0,1\}^{N\times k}:Q\mathbf{1}=\mathbf{1}\: \mathrm{and\:rank}(Q)=k\}$. \\
$\Theta$ & $N\times N$ matrix with $\Theta_{i,j} = \frac{1}{2}\norm{\bm{q}^{(i)}-\bm{q}^{(j)}}^2$ \\
$\odot$ & Hadamard product or element-wise product: $(A \odot B)_{i, j} = (A)_{i, j} \cdummy (B)_{i, j}, \; A, B \in \mathbbm{R}^{m \times n}$ \\
$\norm{Z}_{1,Q}$ & Subspace structured $\ell_{1}$ norm: $\norm{Z}_{1,Q} \stackrel{\cdot}{=} \norm{Z}_1+\alpha\norm{\Theta\odot{Z}}_1=\sum_{i,j}\vert{Z_{i,j}}\vert(1+\frac{\alpha}{2}\norm{\bm{q}^{(i)}-\bm{q}^{(j)}}^2)$, where $\alpha>0$ is trade-off parameter. \\
$S_{\tau} (\nu)$ & $S_{\tau} (\nu) = (| \nu | - \tau)_+ \tmop{sgn} (\nu)$ \\
Affine subspace & A subset $U\subset V$ of a vector space $V$ is an affine subspace if there exists a $u\in U$ such that $U-u=\{x-u|x\in U\}$ is a linear subspace of V. \\
$M$ & \begin{minipage}{0.72\columnwidth} Motion matrix: 
\begin{eqnarray*}{l}
  M = \left[\begin{array}{ccc}
    x_{11} & \ldots & x_{1N}\\
    \vdots &  & \vdots\\
    x_{F1} & \ldots & x_{FN}
  \end{array}\right]_{2 F \times N}
\end{eqnarray*}
where $\{ x_{\tmop{fj}} \in \mathbbm{R}^2 \}^{f = 1, \ldots, F}_{j = 1,
\ldots, N}$ denotes the $2\mathD$ projection of $N$ $3\mathD$ points
on rigidly moving objects onto $F$ frames of a moving camera.\end{minipage} \\
\end{supertabular}

\end{document}

%% file: macros.tex
\usepackage{amsmath,graphicx}
\usepackage{supertabular}
 
\usepackage[numbers]{natbib}
\usepackage{bbm}
\usepackage{bm}
\usepackage{algorithm} 
\usepackage{algorithmic} 
\usepackage{amssymb}
\usepackage{amsthm}
\usepackage{booktabs}
\usepackage{subfig}
\usepackage{longtable}
\usepackage{array}
\usepackage[pdftex]{hyperref}
\usepackage{setspace}	
\usepackage{tikz}
\usetikzlibrary{shapes.geometric, arrows}
\tikzstyle{startstop} = [rectangle, rounded corners, minimum width=3cm, minimum height=1cm,text centered, draw=black, fill=red!30]
\tikzstyle{io} = [trapezium, trapezium left angle=70, trapezium right angle=110, minimum width=3cm, minimum height=1cm, text centered, draw=black, fill=blue!30]
\tikzstyle{process} = [rectangle, minimum width=3cm, minimum height=1cm, text centered, draw=black, fill=orange!30]
\tikzstyle{decision} = [diamond, minimum width=3cm, minimum height=1cm, text centered, draw=black, fill=green!30]
\tikzstyle{arrow} = [thick,->,>=stealth]

\newcommand{\norm}[1]{\left\lVert#1\right\rVert}

\newcolumntype{L}[1]{>{\raggedright\let\newline\\\arraybackslash\hspace{0pt}}m{#1}}
\newcolumntype{C}[1]{>{\centering\let\newline\\\arraybackslash\hspace{0pt}}m{#1}}
\newcolumntype{R}[1]{>{\raggedleft\let\newline\\\arraybackslash\hspace{0pt}}m{#1}}

\catcode`\>=\active \def>{
\fontencoding{T1}\selectfont\symbol{62}\fontencoding{\encodingdefault}}
\newcommand{\Iota}{\mathrm{I}}
\newcommand{\Nu}{\mathrm{N}}
\newcommand{\mathD}{\mathrm{D}}
\newcommand{\tmmathbf}[1]{\ensuremath{\boldsymbol{#1}}}
\newcommand{\tmop}[1]{\ensuremath{\operatorname{#1}}}
\newcommand{\cdummy}{\cdot}
\newcommand{\tmrsup}[1]{\textsuperscript{#1}}
\newcommand{\tmtextit}[1]{{\itshape{#1}}}
